\documentclass{article}
\usepackage[preprint]{spconf2025}
\usepackage{amsmath,amssymb,bm,graphicx}
\usepackage{algorithm}
\usepackage{algpseudocode}
\usepackage{url}
\usepackage{booktabs}
\usepackage{siunitx}
\usepackage{threeparttable}
\usepackage{multirow}
\usepackage{xcolor}
\usepackage[final]{microtype}
\usepackage[colorlinks=true,allcolors=blue,citecolor=blue,linkcolor=blue,urlcolor=blue,breaklinks=true]{hyperref}
\DeclareUrlCommand\repo{\urlstyle{tt}}

\providecommand{\OurOriginalKone}{0.76}
\providecommand{\OurOriginalKoneStd}{0.11}

\providecommand{\OurOriginalKten}{0.92}
\providecommand{\OurOriginalKtenStd}{0.01}

\providecommand{\OurSpheroidjKone}{0.58}
\providecommand{\OurSpheroidjKoneStd}{0.10}
\providecommand{\OurSpheroidjKthree}{0.57}
\providecommand{\OurSpheroidjKthreeStd}{0.12}

\providecommand{\OurSpheroidjKten}{0.70}
\providecommand{\OurSpheroidjKtenStd}{0.08}

\providecommand{\OurDecayKone}{0.66}
\providecommand{\OurDecayKoneStd}{0.05}

\providecommand{\OurDecayKten}{0.80}

\providecommand{\IlastikOriginalKten}{0.85}

\providecommand{\IlastikDecayKten}{0.68}

\providecommand{\DCAMAOriginalKten}{0.94}
\providecommand{\SegGPTOriginalKten}{0.93}

\providecommand{\InHyperBankParams}{306}            

\providecommand{\OurSpheroidjKoneLo}{0.54}
\providecommand{\OurSpheroidjKoneHi}{0.63}
\providecommand{\OurSpheroidjKthreeLo}{0.53}
\providecommand{\OurSpheroidjKthreeHi}{0.62}

\title{HyperBank: A Differentiable Bank of Classical Priors for Few-Shot Spheroid Microscopy Segmentation}

\name{\scalebox{.85}{
\quad M. Průšek $^{1,2}$
\quad A. Novozámský $^{1,*}$
\quad F. Šroubek $^{1}$
\quad T. Volfová $^{3,4}$
\quad V. Svobodová Pavlíčková $^{3}$
\quad S. Rimpelová $^{3,**}$
\thanks{This work was supported by the Czech Science Foundation grant \protect\linebreak GA25-15933S and  CTU student competition grant SGS24/141/OHK4/3T/14. *~Corresponding author for the computer science part (novozamsky@utia.cas.cz). **~Corresponding author for the biology part (silvie.rimpelova@vscht.cz).}%
\thanks{\copyright\,2026 IEEE. Personal use of this material is permitted. Permission from IEEE must be obtained for all other uses, in any current or future media, including reprinting/republishing this material for advertising or promotional purposes, creating new collective works, for resale or redistribution to servers or lists, or reuse of any copyrighted component of this work in other works. Accepted for publication in the IEEE Xplore ICIP~2026 Workshop Proceedings, at the Computational Optical Microscopy Satellite Workshop of the 2026 IEEE International Conference on Image Processing (ICIP), Tampere, Finland.}}}

\address{{\normalsize $^{1}$The Czech Academy of Sciences,  Institute of Information Theory and Automation, Prague, Czechia}  \\        
       {\normalsize $^{2}$Czech Technical University in Prague, Faculty of Nuclear Sciences and Physical Engineering, Prague, Czechia} \\ 
       {\normalsize $^{3}$University of Chemistry and Technology Prague, Department of Biochemistry and Microbiology, Prague, Czechia} \\ 
       {\normalsize $^{4}$Charles University, Faculty of Science, BIOCEV, Vestec, Czechia}}

\copyrightnotice{}
\toappear{To appear in the IEEE ICIP~2026 Workshop Proceedings (Computational Optical Microscopy Workshop).}

\begin{document}
\maketitle

\begin{abstract}
Few-shot spheroid segmentation must adapt to new cell lines, microscopes, and illumination conditions from only a small set of annotated images. While foundation few-shot segmenters can be accurate, their large opaque backbones make it difficult to understand which visual cues drive success or failure. We study this question with HyperBank, a differentiable bank of classical image-processing operators combining Frangi vesselness, a Sauvola threshold pyramid, structure-tensor responses, gradient magnitude, and Laplacian-of-Gaussian filters. HyperBank is fitted on the annotated support images and evaluated on disjoint held-out images across three independently acquired spheroid datasets. We treat it not as a general replacement for foundation models, but as a compact, interpretable few-shot microscopy pipeline and an analytic-prior probe of which classical cues carry the few-shot signal. The results show that, adapted on the same few annotated support images, a compact bank of analytic priors is competitive with --- and on small-cluster, contrast-driven data can outperform --- much larger foundation models, while those models remain stronger on externally sourced, texture-dominated spheroids. Leave-one-family-out ablations indicate that the useful few-shot signal is distributed across operator families and strengthened by support-set-tuned morphology.
\end{abstract}


\section{Introduction}
\label{sec:intro}

Tumour spheroids are three-dimensional multicellular aggregates used as \textit{in vitro} models of tumour growth. Although spheroids are 3D structures, they are usually monitored in 2D bright-field or phase-contrast images, which must be segmented before phenotypes such as area, eccentricity, growth, or necrotic core can be quantified~\cite{SpheroidJ,SpheroScan}. Since each plate may vary in cell line, microscope, illumination, contrast, or morphology, the practical setting is few-shot adaptation from a small annotated support set to the remaining images from the same acquisition regime. Foundation few-shot segmenters such as SegGPT~\cite{seggpt} and DCAMA~\cite{dcama} can achieve strong accuracy but their large pretrained backbones make it difficult to interpret which visual cues drive a particular result. Classical methods such as Sauvola thresholding~\cite{Sauvola}, Otsu thresholding~\cite{globalOtsu}, and pixel-classifier random forests~\cite{ilastik} are more transparent but each encodes only a limited set of assumptions about intensity, texture, or morphology. This motivates a simple question: how far can a compact set of classical image-processing priors go in few-shot spheroid segmentation, and what can it reveal about the visual evidence used across different imaging regimes? To study this, we use HyperBank, a small differentiable bank of analytic operators fitted directly on the annotated support images. Rather than treating the model only as a predictor, we use it as a probe: by removing complete operator families and retraining, we can estimate which priors provide useful evidence and which are partly redundant. The paper makes three contributions: (i) it introduces HyperBank, a compact differentiable few-shot segmentation model based on classical priors; (ii) it evaluates it on three spheroid datasets under a strict shared-split protocol; and (iii) it uses operator-family and post-processing ablations to probe which priors carry the few-shot signal, framing HyperBank as an interpretable microscopy pipeline rather than a replacement for foundation models.

\begin{figure}[b]
\centering
\includegraphics[width=\columnwidth]{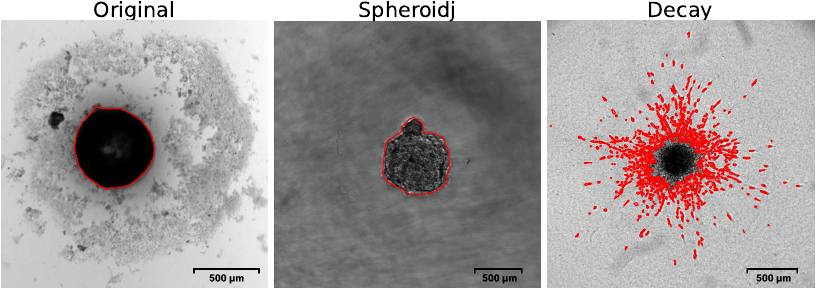}
\caption{A representative sample per dataset, with ground-truth contours overlaid in \textcolor{red}{red}.}
\label{fig:dataset_grid}
\end{figure}


\section{Related Work}
\label{sec:related}

\textbf{Classical spheroid segmentation.}
Spheroid image analysis has a long tradition of compact, inspectable pipelines. Common baselines include global and adaptive thresholding, interactive pixel classifiers, and spheroid-specific tools for measuring projected area, morphology, and growth~\cite{Sauvola,globalOtsu,ilastik,SpheroidJ,SpheroScan}. These methods remain useful in biological screening because they are easy to deploy and their decisions can often be related to simple image properties. At the same time, each method typically relies on a narrow set of assumptions, such as foreground--background contrast, local intensity statistics, or hand-designed pixel features. HyperBank follows this classical line but combines several analytic priors in a single differentiable model fitted from the support set.

\textbf{Differentiable classical operators.}
Several works have shown that classical image-processing operators can be embedded into learnable models. SauvolaNet learns adaptive thresholding for document binarisation~\cite{sauvolanet}, while differentiable Frangi-style filters, Gaussian-derivative layers, and trainable Gabor banks have been used in biomedical and document-image analysis. For example, Wong and Moradi~\cite{wong2022gabor} train a Gabor-kernel bank for fully supervised 3D medical segmentation. Related interpretable pipelines such as Saab and PixelHop~\cite{kuo2016,saab,pixelhop} also replace conventional deep convolutional stacks with structured feature transformations. Compared with these approaches, HyperBank is designed for few-shot microscopy: it keeps the trainable parameter count very small, combines multiple operator families, and uses the fitted bank not only for prediction but also for analyzing which priors contribute to the segmentation.

\textbf{Foundation and SAM-based few-shot segmentation.}
Few-shot segmentation has recently been driven by large pretrained models and dense support--query matching. We therefore compare against representative foundation or SAM-based methods, including DCAMA~\cite{dcama}, SegGPT~\cite{seggpt}, HSNet~\cite{hsnet}, Cellpose-SAM~\cite{cellpose_sam}, $\mu$SAM~\cite{usam}, and a SAM2-based training-free FSS method~\cite{sam2_fss_2025}. These models provide strong practical references but their encoders contain tens to hundreds of millions of parameters and are less direct to interpret at the level of microscopy cues. A related line reduces the number of trainable parameters by adapting frozen foundation backbones with lightweight components, as in PerSAM-F~\cite{persamf}, Matcher~\cite{matcher}, GF-SAM~\cite{gfsam}, FS-MedSAM2~\cite{fsmedsam2}, and biomedical SAM PEFT recipes~\cite{teuber2025peftsam}. HyperBank is orthogonal to this direction: it does not adapt a pretrained backbone but instead tests how far explicit analytic priors can go when fitted directly to the target dataset.

Fully supervised biomedical segmentation methods, including Cellpose3~\cite{cellpose3}, nnU-Net~\cite{nnunet}, and MedSAM~\cite{medsam}, are important reference points but they address a different setting: training or adapting high-capacity models with substantially more supervision or pretrained representation capacity. Few-shot microscopy methods based on large supervised pretraining~\cite{dawoud2020fewshot} are likewise complementary. Our focus is narrower: few-shot adaptation from a small annotated subset of the target dataset, using a compact model with interpretable features grounded in classical image-processing priors.

\section{Methodology}
\label{sec:method}

HyperBank is a compact few-shot segmentation pipeline built around a differentiable bank of classical image-processing operators. Given a small annotated support set from the target dataset, the method fits a small number of operator parameters and a lightweight mixing head, aggregates several independent restarts for stability, and tunes simple binary morphology on the support images. The following subsections describe the feature bank, the fitting procedure, and the support-set-tuned post-processing.

\subsection{Differentiable feature bank}
\label{sec:bank}

The feature bank defines a mapping
$\Phi(x)\in\mathbb{R}^{H\times W\times C}$ from a single-channel image
$x\in\mathbb{R}^{H\times W}$ to a stack of $C{=}32$ per-pixel feature maps. We write the feature stack as
$
\Phi(x)=
\big[
\Phi_{\mathrm{F}},
\Phi_{\mathrm{S}},
\Phi_{\mathrm{ST}},
\Phi_{\mathrm{G}},
\Phi_{\mathrm{LoG}}
\big],
$
where the five blocks contribute $10+10+4+3+5=32$ channels. Each block encodes a different classical cue relevant to spheroid microscopy.

(i)~\textbf{Frangi vesselness}~\cite{frangi}, $\Phi_{\mathrm{F}}$, contributes 10 channels and captures ridge-like or elongated intensity structures at five scales. Each scale emits two channels of opposite ridge polarity --- dark structures on a bright background and bright structures on a dark background --- which share that scale's parameters, so the five scales yield ten feature channels. The per-scale structure-normalisation parameters $\log\gamma\in\mathbb{R}^{5}$ and anisotropy parameters $\log\beta\in\mathbb{R}^{5}$ are learnable, giving 10 trainable parameters.

(ii)~\textbf{Sauvola threshold pyramid}~\cite{Sauvola}, $\Phi_{\mathrm{S}}$, contributes 10 channels and captures local foreground--background contrast by adaptive thresholding at seven window sizes,
$w\in\{15,51,151,400,800,1200,1600\}$ px, spanning cell- to clusters-scale context. The threshold parameter $k_w$ is learnable for each window. Each window produces one differentiable soft-threshold channel, and the three smallest windows additionally produce hard-threshold channels, giving 7 soft and 3 hard Sauvola channels.

(iii)~\textbf{Structure-tensor features}, $\Phi_{\mathrm{ST}}$, contribute 4 channels. They use the smaller and larger eigenvalue magnitudes at two scales and capture local orientation and anisotropy, providing evidence that is less dependent on absolute intensity alone.

(iv)~\textbf{Gradient-magnitude features}, $\Phi_{\mathrm{G}}$, contribute 3 channels. They capture boundary strength at multiple fixed scales and provide direct edge evidence for the spheroid contour.

(v)~\textbf{Laplacian-of-Gaussian filters}, $\Phi_{\mathrm{LoG}}$, contribute 5 channels. They capture multi-scale blob and boundary responses, complementing the gradient channels with second-order local structure.

The resulting feature tensor is normalized with GroupNorm and mapped to a single logit map by a lightweight $3{\times}3$ convolutional head. Only 10 vesselness parameters in $\Phi_{\mathrm{F}}$, 7 threshold parameters in $\Phi_{\mathrm{S}}$, and 289 head parameters ($3{\times}3{\times}32$ kernel weights plus one bias) are trainable, giving \InHyperBankParams\ trainable parameters per restart. The remaining blocks, $\Phi_{\mathrm{ST}}$, $\Phi_{\mathrm{G}}$, and $\Phi_{\mathrm{LoG}}$, are fixed. In the multi-start procedure, this compact fit is repeated four times and the output probabilities are aggregated.

\subsection{Multi-start trimmed-mean fitting}
\label{sec:fit}
For each $K$-image support set, HyperBank is fitted from $M$ perturbed restarts of the same analytic initialization. To reduce overfitting in the smallest-support regimes, the maximum number of epochs is kept lower for small $K$ and increased proportionally with the support-set size\footnote{The maximum number of epochs is set to $18+3(K-1)$.}. Each restart is optimized with Adam using learning rate 0.05, with early stopping after 12 epochs without support-loss improvement. The support loss is a weighted sum of region, instance, and boundary terms:
\begin{equation}
\mathcal{L}
=
\frac{
\mathcal{L}_{\mathrm{BCE}}
+\mathcal{L}_{\mathrm{D}}
+\mathcal{L}_{\mathrm{FT}}
}{2}
+
\mathcal{L}_{\mathrm{CD}}
+
\frac{3}{10}\|\nabla \hat y-\nabla y\|_1 ,
\end{equation}
where $\mathcal{L}_{\mathrm{BCE}}$ denotes binary cross-entropy,
$\mathcal{L}_{\mathrm{D}}$ denotes Dice loss,
$\mathcal{L}_{\mathrm{FT}}$ denotes focal Tversky loss~\cite{abraham2018ftl},
and $\mathcal{L}_{\mathrm{CD}}$ denotes per-component Dice loss~\cite{kofler2023blob}.
The last term is computed on Sobel gradients. At inference, the $M$ fitted models produce probability maps $\hat y_m$, $m\in\{1,\ldots,M\}$. These maps are aggregated pixel-wise by a trimmed mean with trimming rate $\alpha=0.25$:
\begin{equation}
\hat y(i) =
\frac{1}{(1-2\alpha)M}
\sum_{m\in\mathcal{T}_\alpha(i)} \hat y_m(i),
\end{equation}
where $i$ denotes a pixel and $\mathcal{T}_\alpha(i)$ is the set of retained restarts after discarding the $\lfloor\alpha M\rfloor$ largest and smallest probabilities at that pixel. With $M{=}4$, the aggregation removes one high and one low value and averages the two remaining predictions. This keeps the trainable model unchanged while reducing the effect of an outlying restart.

\subsection{Support-set-tuned mask refinement}
\label{sec:morph}
The aggregated probability map is converted to a binary mask by thresholding at 0.5. We then apply a small mask-refinement stage to remove common artefacts: closed holes are filled, binary opening suppresses small protrusions, and connected components below a minimum area are removed. The opening radius $r_{\mathrm{open}}\in\{0,1,2,3,5,7\}$ and the minimum component area $A_{\min}\in\{0,50,200,800,2000,5000,15000\}$ are selected separately for each support set by exhaustive $6\times7$ grid search, maximising IoU on the support images.

\section{Experiments}
\label{sec:experiments}

We evaluated HyperBank in a strict few-shot setting on three spheroid microscopy datasets. The protocol follows the intended deployment scenario: a small number of annotated images from the target dataset was used for adaptation, and performance was measured on a held-out test split from the same acquisition regime.

\textbf{Datasets.}
We used three independently acquired datasets that differed in imaging modality, microscope, magnification, and scene composition (Fig.~\ref{fig:dataset_grid}). The \textsc{Original} is an in-house benchmark of 72 bright-field images of cancer cell spheroids derived from four cell lines: A549 (lung adenocarcinoma), MIA PaCa-2 and BxPC-3 (pancreatic adenocarcinoma), and PC-3 (prostate carcinoma) cells. All images were acquired by an Olympus IX-51 microscope, and each field of view contained one large dark spheroid on a bright background. To keep the dataset sizes matched, we also used 72 images for the other two datasets. The \textsc{SpheroidJ} is a stratified subset\footnote{The exact image identifiers included in this subset are listed at: \\ \url{https://staff.utia.cas.cz/novozada/hyperbank/}} of the public Deep-Tumour-Spheroid bright-field benchmark~\cite{SpheroidJ}. It contains images acquired with two different microscopes, introducing variation in resolution, contrast, and focus. The \textsc{Decay} dataset is an in-house collection of phase-contrast, large-area images containing hundreds of small, partially degraded cells with complex boundaries (microscope Leica DMi8).

\begin{table*}[t]
\centering
\caption{Few-shot IoU for $K\!\in\!\{1,3,5,10\}$ support images, reported as mean$\pm$std over 10 draws. $\dagger$ marks methods not matched to HyperBank's parameter budget.}
\label{tab:main}
\scriptsize
\setlength{\tabcolsep}{1.5pt}
\renewcommand{\arraystretch}{0.95}
\begin{tabular}{l *{12}{c} r}
\toprule
& \multicolumn{4}{c}{\textsc{Original}} & \multicolumn{4}{c}{\textsc{SpheroidJ}} & \multicolumn{4}{c}{\textsc{Decay}} & \\
\cmidrule(lr){2-5}\cmidrule(lr){6-9}\cmidrule(lr){10-13}
Method & $K{=}1$ & $K{=}3$ & $K{=}5$ & $K{=}10$
       & $K{=}1$ & $K{=}3$ & $K{=}5$ & $K{=}10$
       & $K{=}1$ & $K{=}3$ & $K{=}5$ & $K{=}10$
       & Params \\
\midrule
\multicolumn{14}{l}{\emph{Classical (non-DL)}} \\
ilastik-RF
  & $0.60_{\pm.12}$ & $0.79_{\pm.06}$ & $0.83_{\pm.06}$ & $0.85_{\pm.03}$
  & $0.56_{\pm.17}$ & $0.69_{\pm.04}$ & $0.70_{\pm.03}$ & $0.75_{\pm.02}$
  & $0.61_{\pm.11}$ & $0.67_{\pm.05}$ & $0.67_{\pm.04}$ & $0.68_{\pm.02}$
  & $\sim$0.6\,M \\
HyperBank-$\Omega$ (ours)
  & $0.76_{\pm.11}$ & $0.89_{\pm.03}$ & $0.91_{\pm.03}$ & $0.92_{\pm.01}$
  & $0.58_{\pm.10}$ & $0.57_{\pm.12}$ & $0.65_{\pm.11}$ & $0.70_{\pm.08}$
  & $\mathbf{0.66_{\pm.05}}$ & $\mathbf{0.72_{\pm.05}}$ & $\mathbf{0.77_{\pm.01}}$ & $\mathbf{0.80_{\pm.01}}$
  & \textbf{306} \\
\midrule
\multicolumn{14}{l}{\emph{Deep learning}} \\
Cellpose-SAM (ft)
  & $0.69_{\pm.24}$ & $0.91_{\pm.04}$ & $0.93_{\pm.02}$ & $\mathbf{0.94_{\pm.00}}$
  & $0.66_{\pm.18}$ & $0.75_{\pm.10}$ & $0.75_{\pm.07}$ & $0.87_{\pm.02}$
  & $0.37_{\pm.05}$ & $0.46_{\pm.04}$ & $0.49_{\pm.03}$ & $0.51_{\pm.04}$
  & 308\,M \\
SAM-SegHead$^{\dagger}$
  & $0.67_{\pm.23}$ & $0.85_{\pm.04}$ & $0.89_{\pm.05}$ & $0.91_{\pm.03}$
  & $0.44_{\pm.18}$ & $0.54_{\pm.09}$ & $0.63_{\pm.09}$ & $0.70_{\pm.08}$
  & $0.50_{\pm.01}$ & $0.52_{\pm.01}$ & $0.53_{\pm.00}$ & $0.53_{\pm.00}$
  & 90\,K \\
HSNet$^{\dagger}$
  & $0.83_{\pm.06}$ & $0.87_{\pm.04}$ & $0.88_{\pm.01}$ & $0.86_{\pm.01}$
  & $0.77_{\pm.11}$ & $0.84_{\pm.02}$ & $0.84_{\pm.02}$ & $0.84_{\pm.01}$
  & $0.39_{\pm.03}$ & $0.41_{\pm.02}$ & $0.40_{\pm.02}$ & $0.40_{\pm.01}$
  & 25.6\,M \\
DCAMA$^{\dagger}$
  & $\mathbf{0.93_{\pm.01}}$ & $\mathbf{0.93_{\pm.00}}$ & $\mathbf{0.94_{\pm.00}}$ & $\mathbf{0.94_{\pm.00}}$
  & $0.87_{\pm.01}$ & $0.87_{\pm.01}$ & $\mathbf{0.88_{\pm.01}}$ & $0.88_{\pm.01}$
  & $0.49_{\pm.03}$ & $0.51_{\pm.01}$ & $0.51_{\pm.00}$ & $0.51_{\pm.00}$
  & 93\,M \\
SegGPT$^{\dagger}$
  & $0.71_{\pm.25}$ & $0.86_{\pm.17}$ & $0.86_{\pm.17}$ & $0.93_{\pm.02}$
  & $\mathbf{0.89_{\pm.02}}$ & $\mathbf{0.89_{\pm.02}}$ & $\mathbf{0.88_{\pm.02}}$ & $\mathbf{0.89_{\pm.02}}$
  & $0.49_{\pm.00}$ & $0.49_{\pm.02}$ & $0.49_{\pm.01}$ & $0.50_{\pm.00}$
  & 307\,M \\
$\mu$SAM$^{\dagger}$
  & $0.64_{\pm.15}$ & $0.74_{\pm.06}$ & $0.74_{\pm.06}$ & $0.80_{\pm.01}$
  & $0.46_{\pm.14}$ & $0.43_{\pm.12}$ & $0.45_{\pm.09}$ & $0.51_{\pm.05}$
  & $0.23_{\pm.06}$ & $0.23_{\pm.04}$ & $0.23_{\pm.03}$ & $0.23_{\pm.03}$
  & 93\,M \\
SAM2-FSS$^{\dagger}$
  & $0.64_{\pm.08}$ & $0.64_{\pm.06}$ & $0.62_{\pm.03}$ & $0.65_{\pm.01}$
  & $0.47_{\pm.03}$ & $0.46_{\pm.01}$ & $0.46_{\pm.01}$ & $0.45_{\pm.01}$
  & $0.29_{\pm.03}$ & $0.28_{\pm.02}$ & $0.28_{\pm.03}$ & $0.27_{\pm.01}$
  & 80\,M \\
\bottomrule
\end{tabular}
\end{table*}

\textbf{Few-shot protocol.}
For each dataset, we used 48 images as the support pool and a fixed 24-image test split. For each support size $K\in\{1,3,5,10\}$, we sampled $K$ annotated image--mask pairs uniformly from the support pool and evaluated on the test split. This procedure is repeated for 10 independent support draws for each dataset--$K$ combination. All methods were evaluated on exactly the same support/test splits for each draw, and no cross-dataset transfer was used.

\textbf{Baselines.}
All baselines receive the same $K$ annotated support images and are evaluated on the same held-out test images as HyperBank. Each method uses the support set according to its own standard training or few-shot inference recipe; foundation-model features are not inserted into the HyperBank pipeline. The classical reference is ilastik-RF~\cite{ilastik}, using the ilastik 1.4 default 56-channel \textit{fastfilters} stack and a 100-tree Random Forest trained on dense support labels. The deep comparators include Cellpose-SAM~\cite{cellpose_sam}, SAM-SegHead with a frozen MedSAM encoder, HSNet~\cite{hsnet}, DCAMA~\cite{dcama}, SegGPT~\cite{seggpt}, $\mu$SAM~\cite{usam}, and SAM2-FSS~\cite{sam2_fss_2025}. The methods therefore differ not only in capacity but also in how much they adapt to the target dataset: HyperBank and ilastik-RF fit their parameters on the $K$ support images, Cellpose-SAM and SAM-SegHead adapt a lightweight component on top of a frozen encoder, whereas HSNet, DCAMA, SegGPT, $\mu$SAM, and SAM2-FSS perform few-shot inference from generic checkpoints without updating their backbones. We evaluate every method under the same support/test protocol but do not equalise these adaptation budgets, since each reflects the method's intended use. Methods with unmatched parameter budgets are marked by $\dagger$ in Table~\ref{tab:main}; implementation details are provided in the appendix.

\textbf{Metrics and tests.}
We report IoU as mean $\pm$ seed-level standard deviation over 10 support-set samples. Significance is tested with per-image paired Wilcoxon tests and Holm--Bonferroni correction within each (dataset, $K$) setting.

\section{Results}
\label{sec:results}

\textbf{Overall accuracy.}
Table~\ref{tab:main} summarizes the few-shot results across all datasets and support sizes. On \textsc{Original}, HyperBank reached $\OurOriginalKten\!\pm\!\OurOriginalKtenStd$ at $K{=}10$, close to the two strongest foundation-style methods, DCAMA ($\DCAMAOriginalKten$) and SegGPT ($\SegGPTOriginalKten$), while using far fewer parameters than the 307 M-parameter SegGPT encoder. On \textsc{Decay}, HyperBank reached $\OurDecayKten$ at $K{=}10$ and was substantially stronger than all tested foundation-style few-shot methods, which remained near the low-IoU plateau caused by fixed-size token-grid resizing. On the external \textsc{SpheroidJ} set, the foundation-style methods remained strongest. This is expected for this type of data: most images contain a single large compact object, so fixed-grid resizing preserves the relevant object-level structure instead of suppressing many small instances. Overall, the results show that a small analytic feature bank can be highly competitive when segmentation is mainly driven by contrast, boundaries, and cell cluster-scale structure.

\textbf{Comparison with the classical baseline.}
HyperBank was consistently stronger than ilastik-RF on \textsc{Original}. At $K{=}10$, it reached $\OurOriginalKten$ compared with $\IlastikOriginalKten$ for ilastik-RF, with a significant paired Wilcoxon test after correction ($p<10^{-15}$). On \textsc{Decay}, HyperBank also outperformed ilastik-RF for all $K\geq3$; at $K{=}10$, the comparison was $\OurDecayKten$ versus $\IlastikDecayKten$ ($p<10^{-14}$). On the more heterogeneous \textsc{SpheroidJ} dataset, ilastik-RF remained stronger for $K\geq3$. This pattern showed that the analytic bank was not simply a replacement for richer pixel classifiers but a compact and interpretable alternative that works well when the relevant evidence is captured by contrast, boundary, and cell cluster-scale cues.

\begin{figure}[t]
\centering
\includegraphics[width=\columnwidth]{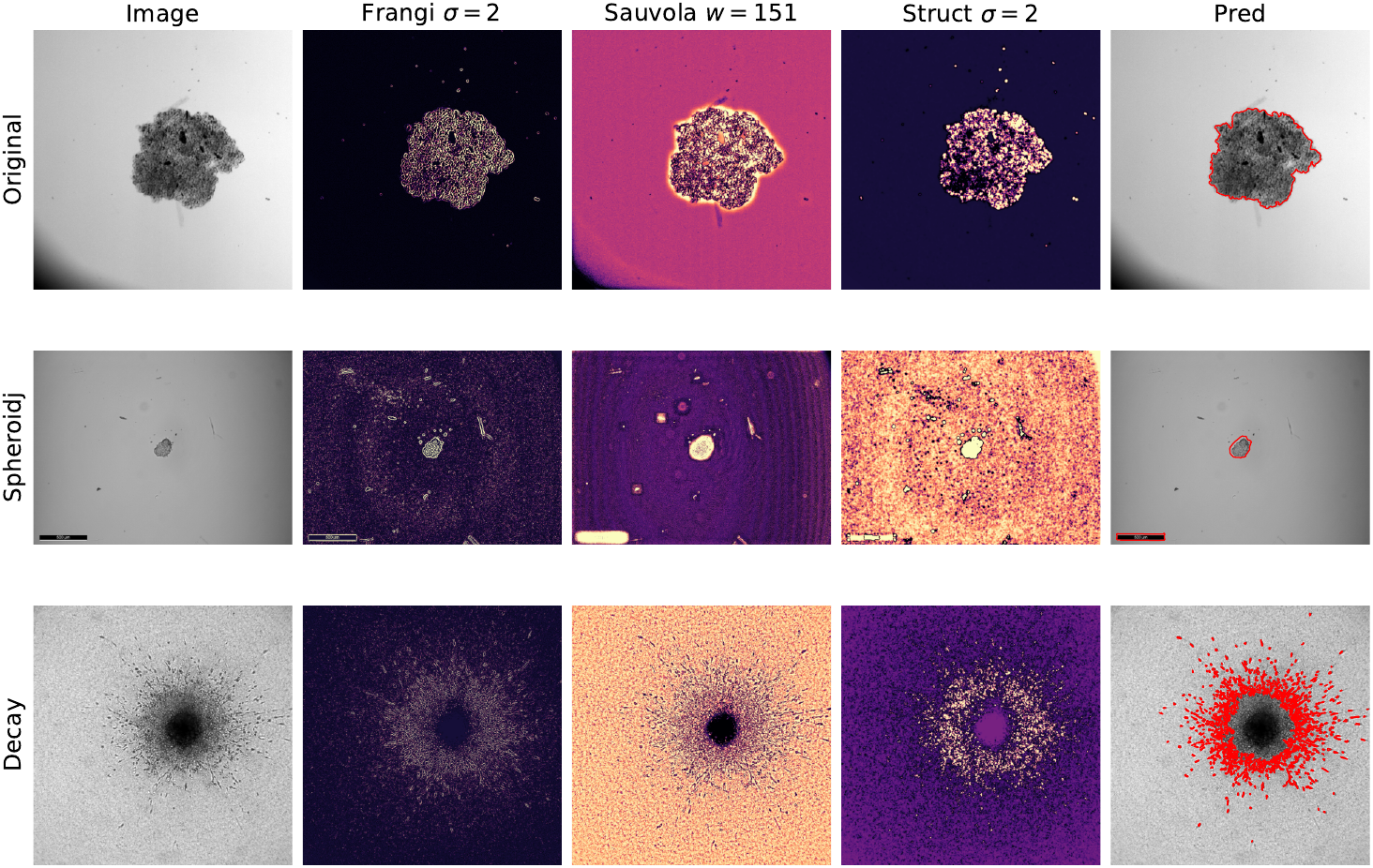}
\caption{Feature maps at $K{=}10$ (seed 0) for one test image per dataset: input, Frangi, Sauvola, structure tensor, and prediction. Feature maps use 2--98\% clipping and a perceptually uniform colormap for display only.}
\label{fig:feature_maps}
\end{figure}

\textbf{The one-shot case.}
At $K{=}1$, HyperBank reached $\OurOriginalKone\!\pm\!\OurOriginalKoneStd$ on \textsc{Original}, $\OurSpheroidjKone\!\pm\!\OurSpheroidjKoneStd$ on \textsc{SpheroidJ}, and $\OurDecayKone\!\pm\!\OurDecayKoneStd$ on \textsc{Decay}, exceeding ilastik-RF in mean IoU on all three datasets. However, performance at $K{=}1$ depended strongly on which image is selected as the only support example. Some support/test combinations remained unreliable. We therefore interpreted the one-shot result as average performance over support draws, not as guaranteed robustness on every image.

\textbf{Ablations.}\label{sec:attribution}
We analyzed three components of HyperBank: the support-set-tuned mask refinement, the feature families, and the size of the mixing head. First, we ablated the mask refinement stage. On \textsc{Original} at $K{=}10$, removing the full refinement reduces IoU by 6.3 pp. Among its individual steps, hole filling had the largest effect (-2.5 pp), followed by the minimum-area filter (-1.7 pp), while opening had little effect on this dataset (-0.2 pp). The entire phase showed a greater decline than any single component, demonstrating that the post-processing steps acted in concert rather than as isolated corrections. Second, we retrained HyperBank after removing one feature family at a time. On \textsc{Original} at $K{=}10$, all single-family removals cost less than 1.5 pp: Frangi (-0.5 pp), Sauvola (-0.7 pp), LoG (-0.9 pp), gradient magnitude (-1.3 pp), and structure tensor (-0.3 pp). This argues against a single decisive handcrafted feature. Instead, the $3{\times}3$ head can reweight the remaining operators when one family is removed. The largest marginal contributor on \textsc{Original} is gradient magnitude; on \textsc{SpheroidJ}, the structure tensor becomes more important (-1.6 pp), consistent with the stronger role of local texture and orientation. Figure~\ref{fig:feature_maps} illustrates representative feature responses from several operator families.

\begin{table}[t]
\centering
\caption{Held-out object statistics. Area is median native object area; Area@1024 is after $1024{\times}1024$ resizing.}
\label{tab:dataset_stats}
\scriptsize
\renewcommand{\arraystretch}{0.95}
\begin{tabular*}{\columnwidth}{@{\extracolsep{\fill}}l c c c c@{}}
\toprule
Dataset & Res. & Obj./img & Area & Area@1024 \\
\midrule
\textsc{Original}  & $1000{\times}1000$ & 1   & 91\,750 & 96\,206 \\
\textsc{SpheroidJ} & $\sim$$1000{\times}1000$ & 1   & 57\,590 & 59\,907 \\
\textsc{Decay}     & $2048{\times}2048$ & $384$ & $161$ & $40$ \\
\bottomrule
\end{tabular*}
\end{table}

\begin{figure}[t]
\centering
\includegraphics[width=\columnwidth]{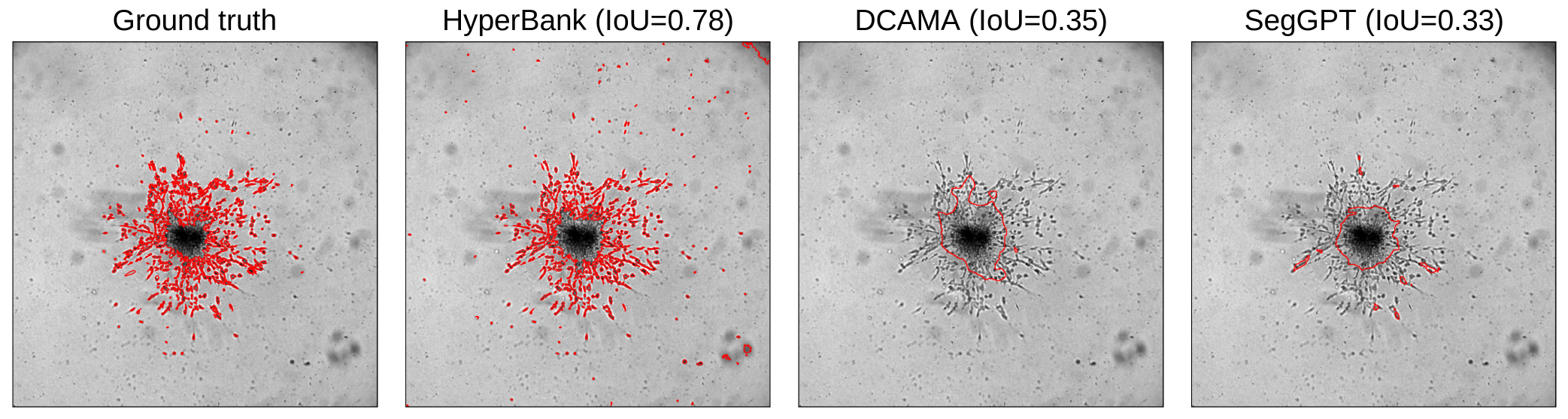}
\caption{Qualitative \textsc{Decay} example at $K{=}10$ (seed 0). HyperBank (IoU $\approx 0.78$) follows individual colonies, while DCAMA and SegGPT collapse to a coarse central region after fixed token-grid resizing.}
\label{fig:qualitative_decay}
\end{figure}

Third, we reduced the mixing head. Replacing the $3{\times}3$ head with a rank-1 separable kernel ($u v^\top$; 56 trainable parameters per restart in total) reduced the cross-mean IoU by about 0.85 pp over \textsc{Original} and \textsc{SpheroidJ} for all $K\in\{1,3,5,10\}$. The largest drops are at $K{=}10$: -1.1 pp on \textsc{Original} and -2.0 pp on \textsc{SpheroidJ}. We kept the $3{\times}3$ head as the main configuration because it gave a measurable accuracy gain while keeping the parameter budget very small.

\textbf{Why fixed-grid foundation models struggle on \textsc{Decay}.}
The main factor is object scale. Unlike \textsc{Original} and \textsc{SpheroidJ}, \textsc{Decay} contains hundreds of small clusters per image (Table~\ref{tab:dataset_stats}). After fixed-grid resizing to $1024{\times}1024$ or $512{\times}512$, many clusters shrink below ViT-patch resolution, so support--query matching captures coarse field appearance rather than individual instances. HyperBank operates at native resolution and preserves these small-cluster cues, outperforming all tested foundation-style few-shot methods at $K{=}10$. Tiled SegGPT inference did not recover the signal (IoU 0.14 vs. 0.50 whole-image), because tiles lose global support--query context (Fig.~\ref{fig:qualitative_decay}).

\textbf{Runtime.}
HyperBank fitting on one NVIDIA L40S GPU took approximately 16 s at $K{=}1$ and 5 min at $K{=}10$, or 25 min for high-resolution \textsc{Decay} at $K{=}10$. Inference took approximately 1.6 s per image with the $M{=}4$ ensemble. Each restart had only \InHyperBankParams\ trainable parameters.

\section{Discussion}
\label{sec:discussion}

\begin{figure}[!t]
\centering
\includegraphics[width=\columnwidth]{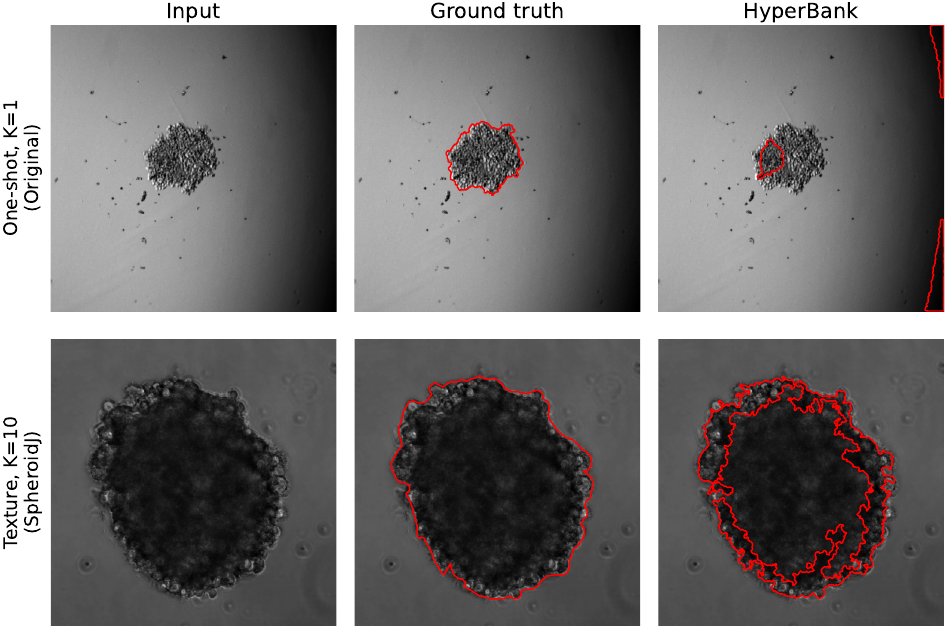}
\caption{Failure cases. \textbf{Top:} one-shot support bias on \textsc{Original}. \textbf{Bottom:} texture-dominated \textsc{SpheroidJ}, where weak boundaries exceed analytic-prior capacity.}
\label{fig:failure}
\end{figure}

HyperBank is a compact, interpretable few-shot microscopy pipeline and an analytic-prior probe, not a general replacement for foundation models. Its purpose is to test how far explicit classical priors can go while keeping the intermediate cues inspectable. The ablations show that no single family dominates on \textsc{Original}: removing Frangi, Sauvola, LoG, gradient magnitude, or structure tensor changes IoU by less than about 1.5 pp, indicating that the useful signal is distributed across partly redundant analytic cues.The results reveal three regimes. On \textsc{Original}, contrast and boundary cues are sufficient for strong performance with very few parameters. On \textsc{Decay}, native-resolution processing preserves many small colonies that fixed-grid foundation-style methods lose during resizing. On the more heterogeneous \textsc{SpheroidJ}, foundation-style methods remain stronger, showing that learned texture and morphology representations are useful when analytic cues are insufficient. Our evaluation already spans bright-field and phase-contrast data, two microscopes in \textsc{SpheroidJ}, and scenes from one large spheroid to hundreds of small colonies. However, fluorescence, inverted contrast, multi-site acquisition, and cross-dataset transfer remain untested. Overall, HyperBank supports a practical workflow: first test whether explicit priors are enough; if not, the ablations indicate where more representation capacity is needed.

\textbf{Failure cases.}
Most low-IoU predictions come from two regimes (Fig.~\ref{fig:failure}). First, in unfavourable one-shot draws, a single atypical support image can bias the fitted thresholds and morphology parameters, causing poor query predictions. This explains the large seed-level spread at $K{=}1$, especially on \textsc{SpheroidJ} ($\OurSpheroidjKone\!\pm\!\OurSpheroidjKoneStd$), and its persistence at $K{=}3$ ($\OurSpheroidjKthree\!\pm\!\OurSpheroidjKthreeStd$). On \textsc{SpheroidJ}, the mean IoU at $K{=}3$ is in fact statistically indistinguishable from $K{=}1$: their 95\% confidence intervals ($\OurSpheroidjKoneLo$--$\OurSpheroidjKoneHi$ and $\OurSpheroidjKthreeLo$--$\OurSpheroidjKthreeHi$) almost fully overlap, so the apparent non-monotonicity reflects support-draw variance on this heterogeneous, two-microscope set rather than a real drop from adding a support image. Second, \textsc{SpheroidJ} remains difficult even at $K{=}10$ ($\OurSpheroidjKten\!\pm\!\OurSpheroidjKtenStd$), because fine intracellular texture and weak diffuse boundaries are not fully captured by contrast-, edge-, and ridge-based priors. In both cases, the support set is either too small or too atypical, or the discriminative information lies in learned texture rather than analytic structure.

\textbf{Reproducibility.}
Code, seeded splits, baseline configurations, and trained mixing-head weights are available at:\\ \small{\url{https://staff.utia.cas.cz/novozada/hyperbank/}}

{\scriptsize
\bibliographystyle{IEEEtran}
\bibliography{refs}}

\appendix

\section{Scope and protocol}

This appendix provides per-baseline implementation details --- checkpoint sources, preprocessing, training/adaptation settings, and inference procedures --- to support reproducibility of the benchmark in the main text. All baselines follow the few-shot protocol of Section~\ref{sec:experiments}: a 48-image support pool, a fixed 24-image test split, $K\in\{1,3,5,10\}$, 10 support draws, and identical support/test splits across all methods. Each method receives the $K$ support pairs through its own fitting or conditioning step and then predicts a binary mask for each test image; the only differences between methods are the internal recipes described below.

\section{Classical baseline}

\subsection{ilastik-RF}

We use ilastik-RF~\cite{ilastik} as the main classical image-analysis baseline. The feature stack follows the default ilastik 1.4 Pixel Classification setup and is computed with the upstream \texttt{fastfilters} library from the \texttt{ilastik-forge} conda channel.

\textbf{Feature stack.}
We use seven scales: $0.3$, $0.7$, $1.0$, $1.6$, $3.5$, $5.0$, and $10.0$. For each scale, we compute Gaussian smoothing, Laplacian of Gaussian, Gaussian gradient magnitude, Difference of Gaussians, the two structure-tensor eigenvalues, and the two Hessian-of-Gaussian eigenvalues. This gives 8 channels per scale and 56 feature channels per pixel in total. We verified on a smoke-test image that the \texttt{fastfilters} output matches the ilastik GUI-level feature output.

\textbf{Classifier.}
We use scikit-learn's Random Forest classifier
(\texttt{RandomForestClassifier}) with 100 trees, Gini impurity, no maximum depth, bootstrap sampling, and class-balanced subsampling. We set \texttt{n\_jobs=-1} and use the experiment seed as the random state.

\textbf{Support labels.}
Although ilastik is often used with sparse scribble annotations, we provide dense ground-truth masks for the $K$ support images. From each support image, we sample 50,000 pixels, balanced between foreground and background whenever possible. If one class contains fewer pixels, the sample is capped by the smaller class. This gives ilastik a strong support signal and makes it a conservative classical baseline.

\textbf{Inference.}
The trained forest predicts a foreground probability for each test-image pixel. We threshold the probability map at 0.5 to obtain the binary mask.

\section{Foundation and SAM-based few-shot methods}

\subsection{HSNet}

We evaluate Hypercorrelation Squeeze Network (HSNet)~\cite{hsnet} using the official PASCAL-$5^i$ split-0 ResNet50 checkpoint. The backbone is ResNet50, and images are resized to $400{\times}400$. No fine-tuning is performed on the support set. For $K{>}1$, we average the $K$ one-shot predictions before thresholding.

\subsection{DCAMA}

We evaluate DCAMA~\cite{dcama} using the official COCO-$20^i$ fold-0 checkpoint with the Swin-B backbone. Images are resized to $384{\times}384$ and ImageNet normalization is applied. The $K$ support pairs are passed to the standard multi-shot prediction routine, and the predicted mask is resized back to the original image size. No backbone or head adaptation is performed.

\subsection{SegGPT}

We use SegGPT~\cite{seggpt} through the HuggingFace checkpoint \repo{BAAI/seggpt-vit-large} and the SegGPT image-segmentation pipeline from \texttt{transformers}. Input resizing is handled internally by the SegGPT processor. For $K{>}1$, we average the $K$ one-shot prediction logits before thresholding.

\subsection{$\mu$SAM}

We evaluate Segment Anything for Microscopy ($\mu$SAM)~\cite{usam} using the light-microscopy generalist checkpoint \texttt{vit\_b\_lm}. Each support mask is converted to a bounding-box prompt, which is then used to segment the query image. The resulting mask is taken as the foreground prediction. SAM's standard fixed-size input preprocessing is applied.

\subsection{SAM2-FSS}

We evaluate the training-free SAM2-based few-shot segmentation pipeline~\cite{sam2_fss_2025}. The method uses augmentative prompting and dynamic matching, as described in the original paper. We run it through an anonymized wrapper around an externally maintained SAM-family one-shot implementation. Inputs are internally resized to the SAM2 token grid.

\subsection{Cellpose-SAM}

We fine-tune the Cellpose-SAM generalist model~\cite{cellpose_sam} on the $K$ support images of each dataset. The model is initialized from the pretrained \texttt{cpsam} checkpoint and trained for 100 epochs with learning rate $5{\times}10^{-5}$, batch size 1, and internal scale \texttt{bsize=256}. Inference uses the standard Cellpose segmentation pipeline. Instance labels are merged into a single binary foreground mask.

\subsection{SAM-SegHead}

For SAM-SegHead, we freeze the MedSAM ViT-B encoder~\cite{medsam} and train a small segmentation head on top of it. The head consists of a $1{\times}1$ channel mixer, two $3{\times}3$ GroupNorm--ReLU smoothing layers, and a final $1{\times}1$ output layer, for a total of 90 K trainable parameters. The MedSAM encoder weights are taken from the HuggingFace repository \repo{flaviagiammarino/medsam-vit-base}. The head is trained for 300 AdamW steps with learning rate $3{\times}10^{-4}$, weight decay $10^{-4}$, and cosine annealing. SAM's standard fixed-size input preprocessing is applied.

\section{Compute and timing}

All methods were run on a single NVIDIA L40S GPU with 48 GB memory, except for ilastik-RF, whose training is CPU-bound and uses 16 logical cores. Table~\ref{tab:timing} reports wall-clock time at $K{=}10$ for one support draw, excluding model loading. For methods that do not train on the support set, the reported ``Train'' time covers support preprocessing and any method-specific conditioning.

\begin{table}[!ht]
\centering
\caption{Wall-clock time at $K{=}10$ on a single NVIDIA L40S GPU. ``Train'' denotes support-set fitting or support preprocessing; ``Infer/img'' denotes average prediction time per test image.}
\label{tab:timing}
\scriptsize
\renewcommand{\arraystretch}{1.0}
\begin{tabular*}{\columnwidth}{@{\extracolsep{\fill}}lcccc@{}}
\toprule
Method & \multicolumn{2}{c}{Train} & \multicolumn{2}{c}{Infer/img} \\
       & $1000^2$ & $2048^2$ & $1000^2$ & $2048^2$ \\
\midrule
\multicolumn{5}{l}{\emph{Classical}} \\
HyperBank & 5.5 min & 31 min & 0.9 s & 6.0 s \\
ilastik-RF & 19 s & 53 s & 1.3 s & 5.7 s \\
\midrule
\multicolumn{5}{l}{\emph{Deep / foundation-style}} \\
Cellpose-SAM & 132 s & 400 s & 0.7 s & 7.6 s \\
SAM-SegHead & 16 s & 31 s & 0.10 s & 0.29 s \\
HSNet & 2 s & 2 s & 0.13 s & 0.30 s \\
DCAMA & 3 s & 4 s & 0.18 s & 0.44 s \\
SegGPT & 5 s & 5 s & 1.78 s & 4.44 s \\
$\mu$SAM & 5 s & 4 s & 1.38 s & 1.42 s \\
SAM2-FSS & 0.3 s & 0.6 s & 4.60 s & 4.76 s \\
\bottomrule
\end{tabular*}
\end{table}

HyperBank is slower to fit on high-resolution \textsc{Decay} images because its feature bank is optimized at native resolution. Reported inference times include the $M{=}4$ ensemble and mask refinement.

\end{document}